\documentclass[letterpaper, 10 pt, conference]{ieeeconf}  

\IEEEoverridecommandlockouts                              

\usepackage{cite}
\usepackage{amsmath,amssymb,amsfonts}
\usepackage{algorithm,algorithmic}
\usepackage{graphicx}
\usepackage{textcomp}
\usepackage{xcolor}

\title{\LARGE \bf
An Investigation of Multi-feature Extraction and Super-resolution with Fast Microphone Arrays
}

\author{Eric T. Chang$^{*1}$, Runsheng Wang$^{*1}$, Peter Ballentine$^{1}$, Jingxi Xu$^{1}$, Trey Smith$^{2}$, Brian Coltin$^{2}$, 
\\ Ioannis Kymissis$^{1}$, and Matei Ciocarlie$^{1}$
\thanks{*equal contribution}
\thanks{This work was supported by a NASA Space Technology Graduate Research Opportunity and by a National Science Foundation Graduate Research Fellowship under Grant No. DGE-2036197.}
\thanks{$^{1}$Columbia University, New York, NY 10027, USA. {\tt\small \{eric.chang@, rw2967@, peter.ballentine@, jxu@cs., johnkym@ee., matei.ciocarlie@\}columbia.edu}}%
\thanks{$^{2}$Intelligent Robotics Group, NASA Ames Research Center, Moffett Field, CA 94035, USA {\tt\small \{brian.coltin, trey.smith\}@nasa.gov}}%
}

\begin{document}

\maketitle
\thispagestyle{empty}
\pagestyle{empty}

\begin{abstract}
In this work, we use MEMS microphones as vibration sensors to simultaneously classify texture and estimate contact position and velocity. Vibration sensors are an important facet of both human and robotic tactile sensing, providing fast detection of contact and onset of slip. Microphones are an attractive option for implementing vibration sensing as they offer a fast response and can be sampled quickly, are affordable, and occupy a very small footprint. Our prototype sensor uses only a sparse array (8-9 mm spacing) of distributed MEMS microphones ($<$\$1, 3.76$\times$2.95$\times$1.10 mm) embedded under an elastomer. We use transformer-based architectures for data analysis, taking advantage of the microphones' high sampling rate to run our models on time-series data as opposed to individual snapshots. This approach allows us to obtain 77.3\% average accuracy on 4-class texture classification (84.2\% when excluding the slowest drag velocity), 1.8 mm mean error on contact localization, and 5.6 mm/s mean error on contact velocity. We show that the learned texture and localization models are robust to varying velocity and generalize to unseen velocities. We also report that our sensor provides fast contact detection, an important advantage of fast transducers. This investigation illustrates the capabilities one can achieve with a MEMS microphone array alone, leaving valuable sensor real estate available for integration with complementary tactile sensing modalities.
\end{abstract}

\section{Introduction}

Designers of robotic tactile sensors can typically choose from many different underlying transduction types, including magnetic, piezoresistive, piezocapacitive, optical,  etc. Among these, microphones are appealing for a number of reasons: off-the-shelf microphone integrated circuits offer fast response time in a low cost package with a very small form factor. In particular, due to the frequency response and fast sampling, microphones effectively respond to vibrations and fast changes in contacts, thus enabling applications such as material identification, fast initial touch detection, and slip detection via vibrations \cite{kappassov2015_review, fishel2012_biotac, fishel2012_biotacinitialtouch, teshigawara2011_pvdf}.

Microphones typically fall in the category of dynamic sensors (which also includes piezoelectrics, accelerometers, etc.), or sensors which respond to \textit{changes} in contact pressures. In contrast, static transduction methods provide a response to constant pressures and deformations. The inability of microphones to provide such static responses means they are generally not used for determining aspects such as contact location or applied force.

A multimodal tactile array could in theory combine multiple transduction types to get the benefit of all worlds, e.g. use microphones for texture classification and static piezocapacitive transducers for contact localization. However, physical space on a sensor's surface is always at a premium, and any new modality adds overhead in terms of the complexity of manufacturing for both the sensor and the signal collection and processing. 

\begin{figure}
\centerline{\includegraphics[width=215pt]{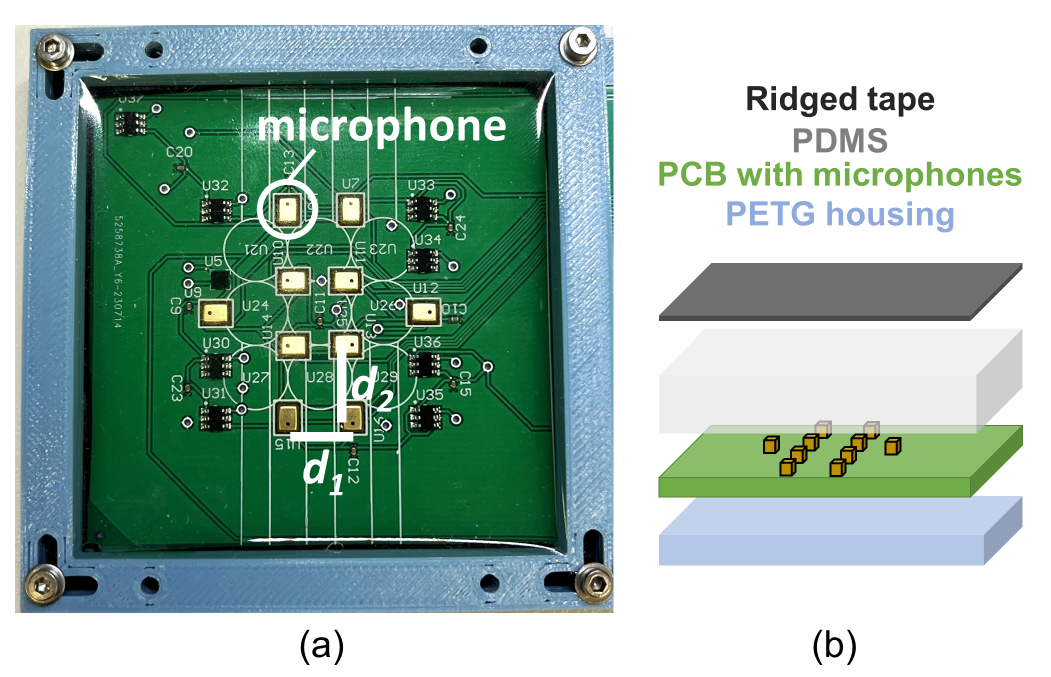}}
\caption{Photo of the microphone array without the textured tape (a), and an exploded view of the sensor fabrication process (b). The inner square of four microphones are spaced $d_1 = 8$ mm apart. The outer ring of microphones are spaced $d_2 = 9$ mm away from the inner square.}
\label{fig:hardware}
\end{figure}

Thus, in this paper, we investigate the possibility of inferring multiple aspects of contact, including texture, location, and drag velocity, using only a distributed array of MEMS microphones as the underlying transducers (Fig.~\ref{fig:hardware}). We aim to combine these abilities with microphones' natural effectiveness in fast detection of initial contact, due to their native high sensitivity and frequency response. We believe these results will prove useful to designers of tactile systems having to decide which underlying transduction modalities to include, how far apart to space individual sensing elements, etc.

A significant development in robotic tactile sensing is the advent of data-driven algorithms that are used on top of the raw sensor signals. For example, learning algorithms have been used with static sensors for object recognition, slip detection, contact localization and super resolution, etc. \cite{calandra2017_feeling, lin2019_learning, piacenza2020_disco}. Likewise, using vibration data from dynamic sensors to learn classifiers for material identification is well studied \cite{fishel2012_biotac, pestell2022_tactip, taunyazov2019_icub, dai2022_cmu, baishya2016_deeplearning, strese2017_multimodalclf}. However, applying learning algorithms to dynamic sensor data to learn other tactile features is underexplored. Here, we use our distributed microphone sensing platform to train models from time domain signals to classify texture, localize contact, and predict drag velocity. In addition, our array also enables fast initial contact detection. Importantly, these are not isolated tasks with carefully selected runtime conditions. Rather, running these models together provides rich, realtime information about the state of sensor contact from solely sparse, dynamic signals. Overall, the main contributions of this paper are as follows:
\begin{itemize}
\item To our best knowledge, we are the first to use distributed microphones to simultaneously classify texture, localize contact, and estimate contact velocity, and show that super-resolution techniques can detect aspects of touch previously unreported using transducers of this type.
\item We combine the capabilities above with the native abilities of dynamic sensors to provide fast response time and high sensitivity for initial contact onset, resulting in a versatile tactile array despite employing a single transducer type.
\end{itemize}

\section{Related Work}
Using vibration-based tactile sensing for material or texture classification has been studied extensively in the robotics literature. Many works use extracted frequency information as input to their classification algorithms \cite{fishel2012_biotac, pestell2022_tactip, taunyazov2019_icub, dai2022_cmu}. More recent work is able to use deep learning algorithms to learn texture classifiers that operate on time series input \cite{baishya2016_deeplearning}. Since there are texture signatures in the frequency domain that are independent of scanning velocity \cite{saal2016_spiketiming, boundy-singer2017_speedinvarianceperception}, it has been shown that texture perception can be impressively robust to different scanning velocities \cite{pestell2022_tactip, strese2017_multimodalclf}. While artificial texture perception has reached impressive performance, these works focus solely on the problem of texture classification. We are interested in what other features can be extracted from a microphone array.

Looking beyond the problem of material classification, there is extensive work on distributed tactile arrays that are sampled at high frequencies and measure vibrations \cite{lin2022_wearable, tomo_uskin, tekscan}. These sensors are typically taxel based, and therefore provide spatial information directly. For example, previous work has taken advantage of the spatial acuity of the iCub's capacitive array \cite{maiolino2013_icub} to combine both spatial and sliding feedback to improve texture classification \cite{taunyazov2019_icub}. While these works focus on sensors that can sense constant pressures and have dense taxel resolution, our study examines sparse dynamic sensors that do not respond to constant pressures.

There are other works on learning from tactile arrays consisting of only dynamic modalities. The NeuTouch \cite{taunyazoz2020_neutouch} and NUSkin \cite{taunyazov2021_nuskin} sensors are state-of-the-art tactile sensors using only dynamic piezoresistive transducers to detect and interpret changes in pressure asynchronously. Using various deep learning techniques designed for spike-based inputs, these sensors have been used for container classification, rotational slip detection, \cite{taunyazoz2020_neutouch} and to detect and decode vibrations traveling through objects \cite{taunyazov2021_nuskin}. However, these works deal with dense dynamic sensing arrays. In this study, we are interested in what features can be learned from a sparser array of microphones.

Lastly, there are a few other works using microphone arrays to learn tactile features. A pair of studies used a tactile skin including sparsely distributed microphones to classify gestures performed by human subjects on the skin via deep learning \cite{yang2021_micskin, yang2023_multiskin}. While the ability to classify human performed gestures is impressive, the work does not explore localization, velocity, or texture models with the sparse microphone array.

\section{Microphone Array Design and Fabrication}
For this work, we built a planar sensor as a test platform for studying the capabilities of distributed MEMS microphones. Our sensor, shown in Fig.~\ref{fig:hardware}, comprises an array of microphones covered with a polydimethylsiloxane (PDMS) elastomer layer, which acts to distribute strain applied at the surface to multiple underlying transducers, mimicking a finger's deformable surface.

In this array, the receptive fields are overlapping: one contact event will activate multiple signals. We did this by adjusting the microphone placement, gel/microphone interface, gel stiffness, and gel thickness. Overlapping receptive fields are valuable when using this platform with high capacity machine learning models that can benefit from multiple rich signal inputs. That said, we limit the microphone density in order to cover a large area while leaving room for other sensors, as future designs might combine a vibratory microphone array with complementary modalities.

\subsection{Electronics}
This sensor contains a printed circuit board (PCB) with a 24x24 mm sensing area. The PCB contains 10 top-port analog MEMS microphones (Knowles SPU0410HR5H-PB, $<$\$1, 3.76$\times$2.95$\times$1.10 mm), spaced 8-9 mm apart (Fig. \ref{fig:hardware}). The microphone signals are fed through a multiplexer, followed by a voltage amplifier with a gain of 5.5 and a bias of 1 V. The microphones are sampled at 2500 Hz with the 12-bit analog-to-digital converter on a Teensy 3.6 microcontroller. This rate is limited by the communication with a PC. We use micro-ROS to communicate between a PC and the Teensy over USB \cite{uROS}.

\subsection{Fabrication}
To fabricate the sensor, we place the PCB in a Prusament polyethylene terephthalate glycol (PETG) housing. We mix PDMS in a 30:1 base to curing agent weight ratio and degas the PDMS in a vacuum chamber to remove all bubbles. We chose a 30:1 ratio as we empirically found that a softer gel (i.e., less curing agent) increased the microphone receptive fields. Under vacuum (-0.75 bar), we pour 17 grams PDMS onto the PCB (6 mm thickness) and remove any remaining bubbles with a heat gun. The vacuum draws the PDMS into the acoustic port, resulting in less hysteresis and increased sensitivity. We cure the sample at 75$^{\circ}$C for 8 hours.

As shown in other work \cite{scheibert2009_fingerprints}, we find implementing a ridged sensor surface important to achieve texture classification. Imitating the papillary ridges (i.e., fingerprints) of the human fingers, the ridges serve to amplify and create vibrations during interaction with other textures. To achieve this, we place a single square of patterned high density polyethylene strips (“Non-Slip Grip Strips” by CatTongue) onto the sensor surface. For consistency, we always use the sensor with the tape applied and replace the tape when the ridges begin to fade. We do not observe any issues of adhesion between the tape and the PDMS. Using a removable tape provides standardized, repeatable friction properties, is more durable than an exposed PDMS surface, and is easily replaceable when it wears down or becomes damaged.

\section{Data-Driven Multi-Feature Extraction}

\begin{figure}
\centerline{\includegraphics[width=150pt]{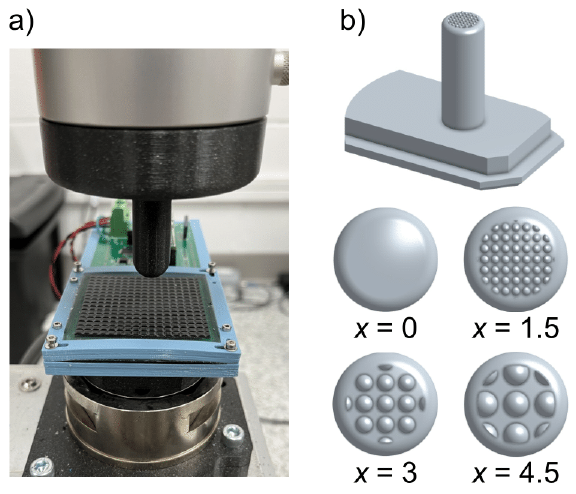}}
\caption{Our tactile sensor on an F/T sensor (a) and the 4 textures used in data collection (b). The four textures (``a," ``b," ``c," ``d") follow $x=$ 0 - 4.5 mm bump spacing and are printed on 15 mm diameter indenters. The bump diameter is determined by the formula $x \sqrt{\frac{2}{\pi}}$ \cite{pestell2022_tactip}. The PLA indenter shown in (a) was used for the response time data (Sec. \ref{sec:responsetime}).}
\label{fig:setup}
\end{figure}

In this study, we leverage machine learning methods to characterize different aspects of contact on our sparse microphone array. Specifically, we use transformer-based architectures to classify texture, localize contact, and estimate contact velocity. The ultimate aim is to investigate the extent to which a microphone array can extract diverse features of a contact episode.

These three tasks are all important elements of tactile perception. Texture discrimination is critical for material classification and tactile exploration. Localization is important for having spatial understanding of a contact. Perceiving contact velocity is an important problem for tasks such as estimating slip velocity and perception during dynamic manipulation scenarios. For slip detection, it is important to estimate velocity over a short time window, since introducing delays removes the ability to use velocity as realtime feedback.

\subsection{Training Data}
Using machine learning models requires a large amount of training data to be collected. In this proof of concept study, we do not move the sensor around to interact with the environment as one might do with a tactile finger. Rather, we imitate contact with various surfaces by dragging different indenters across the sensor. Our collected dataset contains isolated drag episodes with uniquely textured indenters, varying velocities, and varying $z$ forces.

The data collection setup consists of our microphone array mounted onto an ATI Industrial Automation Force/Torque (F/T) sensor (Gamma series, calibration SI-130-10) which provides six signals: force and torque in three axes. This sits below a Universal Robots UR5 robot arm. Our microphone array updates at 2500 Hz, the F/T sensor updates at 2400 Hz, and the robot position and velocity data updates at $\sim$130 Hz. We read the 10 microphones, the 6 F/T signals, and the robot pose and velocity at 2000 Hz.

We use four (``a," ``b," ``c," ``d") different 15 mm diameter textured indenters (Formlabs photopolymer resin), which are mounted onto the UR5 arm. The textures contain hemispherical bumps and are chosen to represent a range of bump densities. The textures are a subset of those used in previous work \cite{pestell2022_tactip}. The bump diameter and spacing is determined in such a way that preserves contact area across the textures (excluding the flat texture), to avoid introducing spurious features unrelated to the texture (Fig. \ref{fig:setup}).

We follow one data collection procedure for all of the training data. We collect drag data at velocities ranging from 20 mm/s to 60 mm/s in 5 mm/s increments using fixed accelerations of 0.3-0.4 m/s$^2$. The $z$ height and $z$ rotation of the indenter are randomized for every drag to be robust to these factors. The height range is calibrated to apply between 1 and 5 Newtons (N) downward force. The rotation range is 0 to 45 degrees to include all indenter pattern rotations. The data collection procedure is detailed in Alg. \ref{alg:data}. In total, we collected $\sim$200 drags per velocity per texture, totalling $\sim$7200 drags.

\begin{algorithm}
\caption{Dragging Data Collection}
\label{alg:data}
\begin{algorithmic}[1]
    \STATE {Calibration: set the $Z$ height bounds ($Z_{max}$, $Z_{min}$) to positions corresponding to 1N and 5N}
    \STATE {Move to 5 mm above a randomly sampled starting point along the border of the sensing area}
    \WHILE{collecting data}
        \STATE{Move to a $Z$ height sampled from [$Z_{min}$, $Z_{max}$]}
        \STATE{Rotate to a $Z$-axis angle sampled from [0, 45] degrees}
        \STATE{Move (while dragging along the sensor) to a new randomly sampled point along the edge of the sensing area that is $\geq$ 15 mm from the current point}
    \ENDWHILE
\end{algorithmic}
\end{algorithm}

\subsection{Preprocessing}
Before training, we do a number of preprocessing steps on the raw data. First, we center the microphone signals around 0 by subtracting a baseline value independently for each microphone. To extract segments of data corresponding to drags, we ignore any data regions where the robot arm was not moving with significant velocity, yielding a collection of segments, each representing a single, complete drag. The dataset is then augmented by sliding a window across each complete drag segment, with overlaps between adjacent windows. 

We empirically adapt window sizes according to the specific task at hand. We use a window size of 100 timesteps (0.05 seconds) for position prediction, 200 timesteps (0.1 seconds) for velocity estimation, and 500 timesteps (0.5 seconds) for texture classification. Across all tasks, we maintain a uniform window offset of 50 timesteps to allow for overlaps. To eliminate the DC components in each window, we apply a third-order bidirectional high-pass filter with a cutoff frequency at 3 Hz. Each resulting sliding window is considered a sample, containing 10-channel sensor readings over $n$ timesteps at a sampling frequency of 2000 Hz. Using this approach, we were able to extract over 110,000 samples from the raw data.

For each sample in our dataset, we extract the texture labels directly from the associated data files. Position labels are acquired by capturing the robot arm’s encoder-reported position at the last timestep within the established sliding window. To obtain velocity labels, we calculate the median velocity within the sliding window from the UR5 velocity encoders, a measure chosen for its robustness to minor fluctuations. 

\begin{figure}
\centerline{\includegraphics[width=200pt]{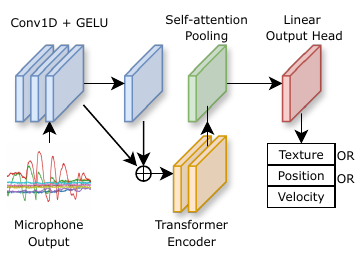}}
\caption{An illustration of our network architecture. We apply 1-D temporal convolution on raw input signals to create latent representations. We feed these representations to a transformer encoder, followed by self-attention pooling and a linear output head.}
\label{fig:arch}
\end{figure}

\subsection{Learning Model Architecture}
Inspired by previous work \cite{zerveas2020transformerbased, baevski2020wav2vec}, we construct a unified signal encoder comprised of a convolution module and a transformer module. The convolution module employs three layers of 1-D strided convolution along the temporal dimension for feature encoding. We apply GELU activation between each pair of convolution layers. The convolutional module outputs a 10-channel latent representation. Before passing the latent representation into the transformer, we apply an additional convolution to the latent representation, pass the output through a GELU activation function, and then sum it with the original latent representation.

We pass the enriched latent representation into a 2-layer transformer encoder block, followed by self-attention pooling \cite{safari2020selfattention}. The pooled output is sent to a classification head for texture discrimination or a regression head for estimating position or velocity. The linear, monolayer output head produces either a texture class via softmax, an $(x,y)$ position vector, or a velocity value, depending on the task.

\subsection{Training}
We employ task-specific methods when partitioning training, validation, and test sets. For texture and position tasks, we assess robustness to changes in drag velocity through a held-out-velocity approach \cite{pestell2022_tactip}. The model is trained on discrete drag velocity profiles in the range of 20 mm/s to 60 mm/s in 5 mm/s increments and evaluated on a distinct velocity profile within the range but not in the training set. For example, when testing classification accuracy on the 40 mm/s test set, we use a model trained on all velocity profiles except 40 mm/s. We use 5 different velocity profiles as the held-out set (i.e., we train 5 different models).

For the velocity task, we train on all velocity profiles and execute 10-fold cross-validation, each utilizing a unique held-out test set randomly sampled from all velocity profiles. Whenever sampling from the datasets, we sample individual drags before dividing into windows to ensure that adjacent windows from the same drag do not end up in different training/validation/test sets.

For classification and regression tasks, we use cross-entropy and mean squared error (MSE) loss, respectively. We select the best-performing model using validation loss, and report test set performances using the selected model. At inference time, we run the three models in parallel.

\section{Contact Characterization Performance}

\subsection{Texture Classification}
\begin{figure*}[!t]
\centerline{\includegraphics[width=\textwidth]{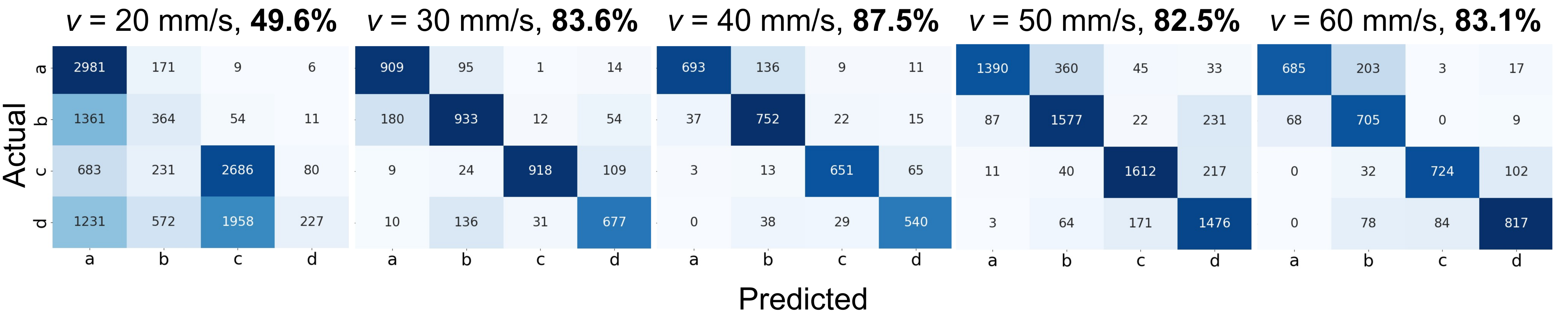}}
\caption{Confusion matrices on 4 class texture classification. Each matrix is for a different test set, each of which correspond to the held-out velocities.}
\label{fig:texture}
\end{figure*}

Upon training a texture classifier, we evaluate the model’s ability to distinguish between the four textures while generalizing to velocities not seen in training. We plot confusion matrices across five different test sets, where each matrix is the performance on a velocity that was held out during training (Fig. \ref{fig:texture}). The average accuracy across the 5 test sets is 77.3\%. The model’s robustness to varying velocities and strong performance on unseen velocities suggests that the model learns to classify textures based on signal features that are independent of velocity.

We note that the 20 mm/s velocity has significantly worse performance than the other velocities. This is likely due to a slower drag causing less prominent signals, and therefore fewer distinguishing characteristics between the textures. Poor classification performance on this velocity relative to higher velocities is consistent with previous work \cite{pestell2022_tactip}. Excluding this velocity, the average classification performance among the other test sets is 84.2\%. We also observe from the confusion matrices that when samples are misclassified, the models tend to predict a neighboring texture.

\subsection{Localization}
The localization regressor estimates the current contact position from a history of data (100 timesteps, 0.05 seconds). As with the texture model, we examine the localizer’s performance on five test sets containing unseen velocities. The mean error is 1.8mm. 

We find this learned position model to outperform a heuristic-based method, which assigns the position of the microphone with the highest signal-to-noise ratio (SNR) over the history window as the estimated location. The mean / median localization error for the baseline is 5.5 / 5.3 mm. We note that using SNR over the history window means that this baseline predicts the average position over the entire window of history, whereas the learned model is trained to estimate the most recent position. That said, the delay introduced by such a short window (0.05 seconds) is small.

\subsection{Direct Velocity Regression}

We analyze velocity estimation results when the model has access to different windows of history: 0.05, 0.10, and 0.25 seconds. We achieve a mean absolute velocity error of 6.5, 5.6, and 5.7 mm/s for 0.05, 0.10, and 0.25 second history windows, respectively, or 16.3\%, 14.0\%, and 14.3\% of the 20 - 60 mm/s range of velocities explored (Tab. \ref{tab:results}). We find that a window larger than 0.05 seconds of history is necessary for learning meaningful velocity estimation; however, results do not noticeably improve for the 0.25 s window (Tab. \ref{tab:results}). 

We hypothesize that directly estimating velocity with a designated model will be more accurate than using the derivative of a position measurement. To test this hypothesis, we compare our results to inferring velocity using the learned position model. We make this velocity estimate by using the estimated start and end positions over a 200 timestep (0.10 s) time window, resulting in a median error of 9.7 mm/s. It is not surprising that the learned model outperforms this baseline, as errors in the position estimation will be amplified when computing velocity over a short time window. However, for the conditions investigated, this result suggests that there is value to making direct velocity estimates.

\begin{table*}[]
\renewcommand{\arraystretch}{1.4}
\caption{Localization and velocity estimation error. Localization performance is reported on velocities not seen in training (overall mean error is 1.8mm). Velocity estimation is trained on all velocities and error is reported in 5mm/s velocity bins and for different history lengths (overall mean velocity error is 6.5 / 5.6 / 5.7 mm/s for a 0.05 / 0.10 / 0.25 s history window).}
\label{tab:results}
\centering
\begin{tabular}{c|c||c|c|c|c|c|c|c|c|c}

\multicolumn{2}{l}{}             &  \multicolumn{9}{c}{Drag Speed (mm/s)}      \\
\hline
Task                             & History (s) & 20          & 25          & 30          & 35          & 40          & 45          & 50          & 55          & 60 \\
\hline\hline
M/Mdn Loc. Err. (mm)             & 0.05        &  2.0 / 1.6  &  -          &  1.9 / 1.6  &  -          &  1.6 / 1.3  &  -          &  1.8 / 1.6  &  -          &  1.8 / 1.5  \\
\hline
M/Mdn Vel. Err. (mm/s)           & 0.05        &  7.9 / 6.4  &  5.5 / 4.5  &  4.3 / 2.8  &  5.5 / 5.1  &  6.0 / 5.6  &  6.3 / 5.2  &  6.3 / 4.5  &  7.0 / 4.8  &  9.7 / 7.8  \\
M/Mdn Vel. Err. (mm/s)           & 0.10        &  6.3 / 5.5  &  5.0 / 4.3  &  4.2 / 2.9  &  5.0 / 4.5  &  5.1 / 4.4  &  5.1 / 4.0  &  5.2 / 3.6  &  6.1 / 4.4  &  8.8 / 7.3 \\
M/Mdn Vel. Err. (mm/s)           & 0.25        &  9.7 / 9.5  &  8.0 / 8.0  &  5.1 / 4.3  &  4.8 / 3.8  &  5.4 / 5.0  &  4.4 / 3.7  &  3.7 / 2.7  &  3.9 / 2.7  &  6.6 / 5.7  \\
\hline

\end{tabular}
\end{table*}

\section{Response Time Analysis}
\label{sec:responsetime}
While being able to extract diverse features from a microphone array is useful, the primary benefit of using vibration modalities is for their fast response time and sensitive contact detection. Therefore, to complement our machine learning study, we also provide an analysis of this sensor’s response time to illustrate this benefit in a MEMS microphone array.

To achieve this, we measure the relative response time between a single microphone and an F/T sensor used as a baseline – i.e., the difference between the time the F/T sensor and the microphone detected contact. While an F/T sensor is not a ground truth measurement of the start of contact, we treat it as a baseline as follows: we use the F/T data offline to estimate when the true contact began (i.e., we look ahead in the data to assign contact labels), resulting in contact identification while the F/T signal is still in the sensor's noise range -- earlier than the F/T sensor could have detected it in realtime. We detail this procedure below.

\subsection{Data Collection and Processing}
To collect data for this experiment, we use the same data collection setup as for the training data, but with a hemispherical, polylactic acid (PLA) indenter on the UR5 arm, and with a downward motion instead of a dragging one (Fig. \ref{fig:setup}). We also increase the rate at which the F/T sensor updates the ethernet port to 5000 Hz. However, we note that the internal F/T strain gauge measurements always occur at 7000 Hz \cite{atidatasheet}. We also increase the rate at which data is synchronized to a file to 2300 Hz. We selected one microphone with a representative sensitivity for this analysis.

The data collection process consisted of an indentation procedure repeated over 45 episodes. The procedure consisted of: 1) move the indenter above the sensor such that a constant velocity can be reached before contact, 2) move the indenter downwards at a preset velocity (acceleration = 4.0 m/s$^2$) onto a preset location at a depth correlated with 2-3 N downward force, and remain on the sensor for 1 second.

We select the contact points to be 0, 2, 4, and 6 mm away from the microphone of interest, and we set the incoming velocities to be 10, 55, and 100 mm/s. Empirically, we find the microphone receptive fields to significantly decay around 4 mm away from the microphone, so we include 6 mm in our analysis to look at the response beyond the nominal receptive field. In practice, 6mm away from a single microphone is always well into the receptive field of another (Fig. \ref{fig:hardware}).

Before analyzing where contact starts, we first use a section of data at the beginning of each episode to individually bias (i.e., tare) the episode to offset drift in the F/T readings. We also remove any episodes in which the contact data was affected by the F/T sensor momentarily flat-lining and failing to update. After these steps, we are left with at least 30 episodes per location and velocity condition.

\textbf{Identifying F/T Contact Time.} To identify the time at which the F/T sensor detects contact for each episode, we use an offline heuristic to identify the start of contact as early as possible. The heuristic would not work in realtime because it looks both forward and backward in the data. By sliding a window function across each episode, we found this combination of heuristics to identify contact consistently (treating downward forces as negative): 1) at least 1 point in the window is less than -1 N; 2) all remaining points in the window are less than the current point; 3) the curve starts decreasing within the next 4 timesteps; and 4) the curve decreases significantly 25 timesteps later. We verify that each episode was processed correctly, noting that the heuristic is not perfect and sometimes identifies contact too early, particularly for the (0mm, 55 mm/s) dataset.

\textbf{Identifying Microphone Contact Time.} Whereas an offline heuristic is appropriate for the F/T data, identifying the microphone contact time requires a realtime heuristic so that the measured response time is relevant in practice. We chose a heuristic that identifies contact when the microphone signal is at least 18 ADC counts higher than the median of the last 20 signals. Intuitively, this examines if the signal has sufficiently exceeded the noise floor. We found this heuristic to work in realtime with almost no false positives.

\subsection{Response Time Analysis Results}

\begin{table}[]
\renewcommand{\arraystretch}{1.3}
\caption{Mean (standard deviation) relative response time (in ms) between the microphone and the F/T sensor. }
\label{tab:responsetime}
\centering
\begin{tabular}{c|c|c|c|c}
                               & 0 mm      & 2 mm      & 4 mm      & 6 mm      \\ \hline\hline
\multicolumn{1}{c||}{10 mm/s}  & 3.0 (3.1) & 4.2 (2.5) & 5.1 (1.4) & -         \\ \hline
\multicolumn{1}{c||}{55 mm/s}  & 4.8 (2.5) & 3.1 (1.4) & 3.0 (1.3) & 4.0 (1.6) \\ \hline
\multicolumn{1}{c||}{100 mm/s} & 3.0 (1.3) & 2.6 (1.7) & 3.1 (1.6) & 3.8 (2.3) \\ \hline
\end{tabular}
\end{table}

We find the average response time of the sensor to range from 2.6 - 5.1 ms across the conditions studied (Tab. \ref{tab:responsetime}). As one would expect, the response time tends to increase as the tap moves further from the microphone. At the (6 mm, 10 mm/s) condition, the contacts were not detected using our microphone contact heuristic. That said, these responses have signal changes noticeable to the naked eye, but they are small relative to the other responses and come as far as 200 ms after initial contact, making them impractical to use in realtime.

A 3 ms response correlates to a response above 330 Hz -- faster than many static sensor arrays. However, the microphone has a reported sensitivity limit of 10 kHz and we sample the system at 2300 Hz, putting the Nyquist frequency at 1150 Hz – far higher than our measured response time. We therefore attribute the limiting factor of the response time to the latency introduced by the elastomer.

\textbf{Limitations of this Analysis.} The F/T sensor is not a precise ground truth measurement for the beginning of contact as it has its own response characteristics. The frequency response's -3dB cutoff is around 1800 Hz, and it can deliver data to the ethernet port with a 0.286 ms delay~\cite{atidatasheet}. In addition, we do not quantify delays from Ethernet port reading, Teensy serial communication, or ROS2. Ultimately, this experiment provides an estimate of the response time relative to a fast or quasi-real-time baseline.

\section{Conclusions}

This study has explored the ability of MEMS microphone arrays to provide information about multiple aspects of contact including simultaneous texture classification, contact localization, and contact velocity determination. Contact localization and velocity determination are previously unreported using this transduction modality. We attribute the ability to determine such features to data-driven techniques that make use of short windows of data as opposed to single snapshots, processed by a self-attentive transformer architecture that is known to be highly effective at analyzing time-series data. Additionally, we highlight MEMS microphones' capacity for rapid initial contact detection using simple heuristics. 

We also discuss two primary limitations of our work with microphone arrays. Unlike some piezoresistive or piezoelectric arrays, microphones must be placed on a PCB and are difficult to arrange in a dense taxel grid. In addition, this work's tactile feature extraction relies on learned models, which may degrade when exposed to stimuli outside of the training distribution. This can become apparent if the signal distribution shifts with sensor use.


This study showcases that a diverse set of tactile features can be extracted from \textit{only} a sparse microphone array. Thus, using microphones in this way can be a valuable method of implementing fast, low cost, and space-efficient vibration sensing. In the pursuit of information rich tactile sensors that mimic both the static and vibration receptors in human hands, we hope these results will provide insight into the capabilities of MEMS microphones as a tactile modality.


\addtolength{\textheight}{-5cm}   

\bibliographystyle{IEEEtran}
\bibliography{bib/main}

\end{document}